\documentclass[conference]{IEEEtran}
\IEEEoverridecommandlockouts
\usepackage{cite}
\usepackage{amsmath,amssymb,amsfonts}
\usepackage{algorithmic}
\usepackage{graphics}
\usepackage{lipsum}
\usepackage{grffile}
\usepackage{hyperref}
 \usepackage{textcomp}
\usepackage{footmisc}
\usepackage{pdfpages}
\usepackage{ifpdf}
\usepackage{epstopdf}
\usepackage{footnote}
\usepackage{url}
\usepackage{blindtext}
\usepackage{xcolor}
\usepackage[utf8]{inputenc}
\usepackage[bottom=0.95in,top=0.7in]{geometry}
\newcommand{\CLASSINPUTbottomtextmargin}{0.95in}
\usepackage{babel,textcomp}
\usepackage{subfigure}
\usepackage{csquotes}
\def\BibTeX{{\rm B\kern-.05em{\sc i\kern-.025em b}\kern-.08em
    T\kern-.1667em\lower.7ex\hbox{E}\kern-.125emX}}

\begin{document}
\title{BERT-based Ensemble Approaches for Hate Speech Detection}
\author{\IEEEauthorblockN{Khouloud Mnassri, Praboda Rajapaksha, Reza Farahbakhsh, Noel Crespi}
\IEEEauthorblockA{\textit{ Telecom SudParis, IMT, Institut Polytechnique de Paris,}
91764 Palaiseau, France\\
\{khouloud.mnassri, praboda.rajapaksha, reza.farahbakhsh, noel.crespi\}@telecom-sudparis.eu}}

\maketitle

\begin{abstract}
With the freedom of communication provided in online social media, hate speech has increasingly generated. This leads to cyber conflicts affecting social life at the individual and national levels. 
As a result, hateful content classification is becoming increasingly demanded for filtering hate content before being sent to the social networks. This paper focuses on classifying hate speech in social media using multiple deep models that are implemented by integrating recent transformer-based language models such as BERT, and neural networks. 
To improve the classification performances, we evaluated with several ensemble techniques, including soft voting, maximum value, hard voting and stacking. 
 We used three publicly available Twitter datasets (Davidson, HatEval2019, OLID) that are generated to identify offensive languages. We fused all these datasets to generate a single dataset (DHO dataset), which is more balanced across different labels, to perform multi-label classification. Our experiments have been held on Davidson dataset and the DHO corpora. The later gave the best overall results, especially F1 macro score, even it required more resources (time execution and memory).
 The experiments have shown good results especially the ensemble models, where stacking gave F1 score of 97\% on Davidson dataset and aggregating ensembles 77\% on the DHO dataset.
\end{abstract}


\section{introduction}
\vspace{-5pt}
Twitter is used to disseminate hate speech, especially with the anonymity provided \cite{anonym}.
Thus, any strategy to identify such content is critical in today's world for keeping the internet a safe environment.
Detecting online hateful content is the first step in developing a system that flags such items and take right actions.
Human annotators are employed by social media corporations to erase these samples and users can flag anything they find harmful to the public. But these procedures are time consuming and depend on human
judgment. Thus, automated hate speech detection approaches have been a major concern in this era. 
To this end, in early research, many attempts were built upon machine learning algorithms and different features extraction techniques, but in recent years, significant performances were obtained by using Transformer-based Language Models \cite{transformer}.
\vspace{-2pt}
Bidirectional Encoder Representations from Transformers, BERT, has achieved state-of-the-art results in many NLP tasks \cite{bert}. Moreover, Neural Network - NN approaches managed to reduce feature engineering and are implemented and coupled with the representation of texts as word vectors through word embedding models. 
However, machine learning and NN models required to have a larger corpus of datasets to train, but BERT-based models work with a small number of labeled data and sometimes. Hence, to use the advantage of both approaches, we integrate them together to build a deep architecture in order to achieve considerable performances in hate speech classification. In addition, we use Ensemble learning approaches to improve model performances further \cite{ensemble}.
In this context, we employed and fine tuned BERT, combined it with deep neural networks and merged obtained models via ensemble learning, in order to detect hate speech in Twitter base data.

The main contributions of this paper are:
1) Presenting transfer learning approach by integrating BERT with Multi Layers Perceptron (MLP), Convolutional Neural Networks (CNN) and Long short-term memory (LSTM) for hate speech detection, in order to explore this mix of models performs better in text classification tasks compared to neural networks or BERT alone. 2) Applying several ensemble learning approaches to improve the performance of these models. 3) Creating new corpora, by fusing multiple public and labeled datasets to get large-size and more balanced corpora. 4) Comparing the integrated models with baseline models and, 5) Comparing model performances through their memory utilization and runtime parameters. 
 We managed to get benefit from the contribution of the combination of BERT and deep neural networks in the text classification task, dealing with the scarcity and the imbalanced data. In addition, we exploited that ensemble learning approaches enhanced our deep models' ability in the classification of hate speech content.
\vspace{-7pt}

\section{Literature Survey}
\textbf{Transfer learning:}
The deep learning models rely on the adoption of Neural Networks - NN coupled with conventional word embedding techniques, which achieved effective performance but often not as efficient as a transformer's \cite{transf}.
Recent methodologies have gradually changed approaches from RNNs to self-attention and transformers \cite{self} in many NLP tasks.
Google’s BERT \cite{bert} adapted Transformers in 2018, that can condition both left and right context to pretrain deep bidirectional representations from texts. 
Chiril et al \cite{chiril} built a BERT-based muti-task hate speech detection method that outperforms a system trained on a single topic-generic dataset.
Moreover, Kovács et ai \cite{kovacs} suggested a model of a conjunction of RoBERTa and FastText incorporated with CNNs and RNNs and achieved 63\% F1 score.
Malik et al. \cite{davidsonSOTA} conducted a review of 14 shallow/deep classifiers, driven by a variety of word representation approaches. They resulted that coupling BERT, ELECTRA, and Al-BERT with NNs outperforms other approaches. Their best models were BERT+CNN and ELECTRA+MLP giving F1 macro score of 76\% on Davidson \cite{davidson} dataset.
Moreover, Mozafari et al. \cite{mozafari} used BERT-based methods (BERT+Bi-LSTM, and BERT+CNN) to detect hate speech and achieved significant performances.

\textbf{Ensemble learning:}
This is a machine learning approach that involves training several models together in a given task.
An ensemble is made up of a group of learners known as base/weak learners, trained for the same problem, then integrated to improve results \cite{ensemble}.

Badjatiya et al. \cite{badjatiya} ensembled Embedding, LSTM and gradient boosted trees to determine whether a tweet is racist, sexist, or neither in a 16k dataset and achieved 93\% F1 score.
Plaza-del et al. \cite{arco} built a vote ensemble classifier including SVM, LR and Decision Tree(DT) to classify hate tweets in English and Spanish and achieved 44\% F1 score. Agarwal et al. \cite{agarwal} proposed a Stacking classifier to extract word embedding with RNNs, then, they used SVM, DT, MLP, kNeighbors, ELM, with LR as the meta-classifier and F1 score of 73\%. 
Aljero et al. \cite{aljero} got F1 score of 97\% with their stacking ensemble that combines SVM, LR, and XGBoost. 

According to the recent researches that have been held using ensemble learning, we find that the majority of them didn’t implement transformers in their experiments (whether for word embedding or as classifiers).
Moreover, most of the recent studies were focused on assembling only machine learning classifiers.
The classification process was also binary detection of hateful/abusive content.\\
Based on the state-of-the-art on hate speech detection using transformers-based transfer learning and ensemble learning, in this work we propose several strategies to the same task by integrating best of both approaches. 

\begin{figure} 
\centering
\includegraphics[width=2.5in]{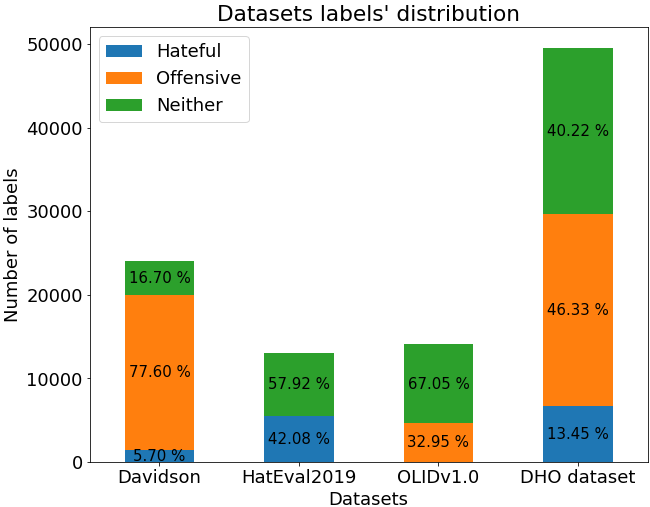}
 \caption[my images]{Datasets labels distribution.}
\label{data}
\vspace{-7pt}
\end{figure}

\vspace{2pt}

\section{Methodology}
We aim to build deep learning hate speech detectors by fine tuning and integrating BERT with several NNs, and to enhance hate speech detection performance via ensemble learning. We work on multi-label classification task to distinguish between hateful, offensive and normal content from Twitter.
\vspace{-12pt}
\subsection{Dataset}
\vspace{-5pt}
Our analyses are based on 3 publicly available datasets: \textbf{Davidson} \cite{davidson}, \textbf{HatEval} \cite{hateval},
and \textbf{OLID}  \cite{olid}.
{\textbf{1) Davidson:}} created by Davidson et al.\cite{davidson} (2017).  
It includes 24783 tweets and 3 labels (hateful, offensive, and neither), that were generated using Figure Eight crowdsourcing\footnote{\url{https://appen.com/figure-eight-is-now-appen/}\label{this}}.
These tweets were selected from 85.4M archive tweets, focusing on HateBase keywords (hatebase.org), and annotated by 3 people. For the rest of the paper, we refer this dataset as \textit{Davidson} dataset.
{\textbf{2) HatEval:}}
contains 13000 tweets about immigrants and women, and generated for the SemEval2019 Task5\cite{hateval}. The majority of the tweets came from the AMI corpus\footnote{\url{https://groups.inf.ed.ac.uk/ami/corpus/}\label{}} and the dataset was labeled via Figure Eight crowdsourcing\footref{this}. 

The annotators detect if a tweet is hateful, aggressive, and whom it is directed (individual or group). In this paper, we refer this dataset as \textit{HatEval2019}.
{\textbf{3) OLID:}} Offensive Language Identification Dataset, created by Zampieri et al. \cite{olid} and composed of 14100 tweets.
A 3-level hierarchical annotation was used, with the first one determines if a tweet is offensive. In this paper, We refer this dataset as \textit{OLID} dataset.
\textbf{4) DHO:}
generated by merging 
\textbf{D}avidson, \textbf{H}atEval2019 and \textbf{O}LID datasets to build a large data corpus for our analysis. 
Data statistics of these corpora are illustrated in Figure \ref{data}.
Compared to \textit{Davidson}s, \textit{DHO} is more balanced, it is used to demonstrate the resilience and generalization of our models.
\vspace{-4pt}
\begin{figure} 
\centering
\includegraphics[width=2.2in]{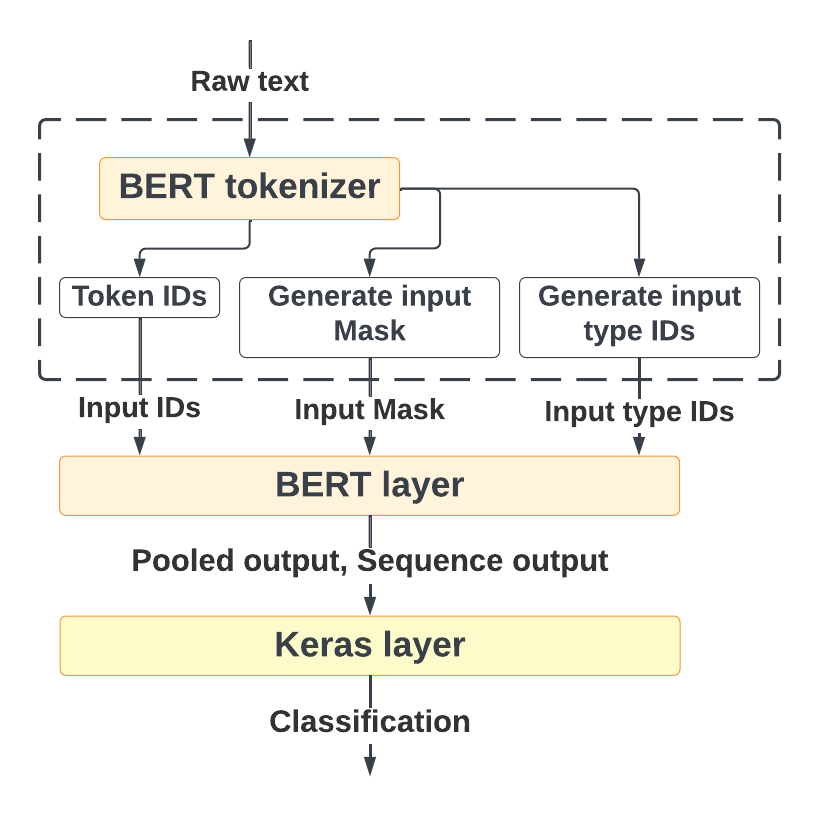}
 \vspace{-1.5\baselineskip}
\caption[my image]{General Architecture: BERT+Deep Neural Networks.}
\label{bert++}
\vspace{-7pt}
\end{figure}
\begin{figure*}
    \centering
    \subfigure a.{\includegraphics[width=0.22\textwidth]{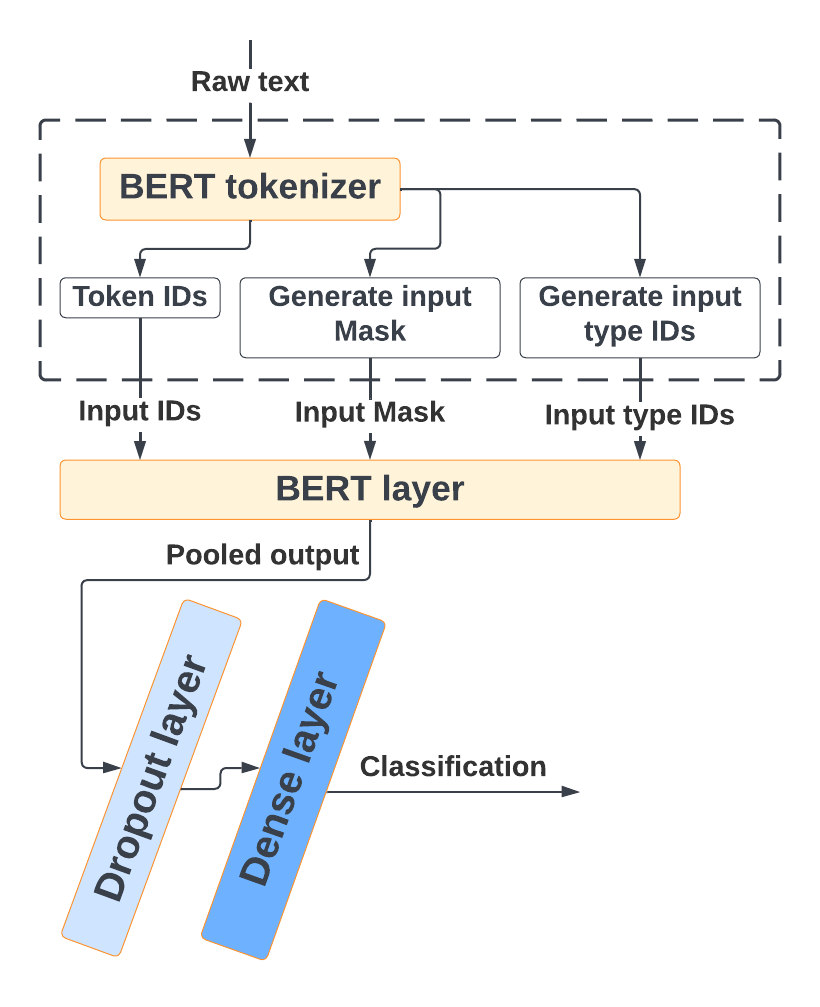}}\label{a}
    \hspace{-1.5\baselineskip}
    \subfigure b.{\includegraphics[width=0.22\textwidth]{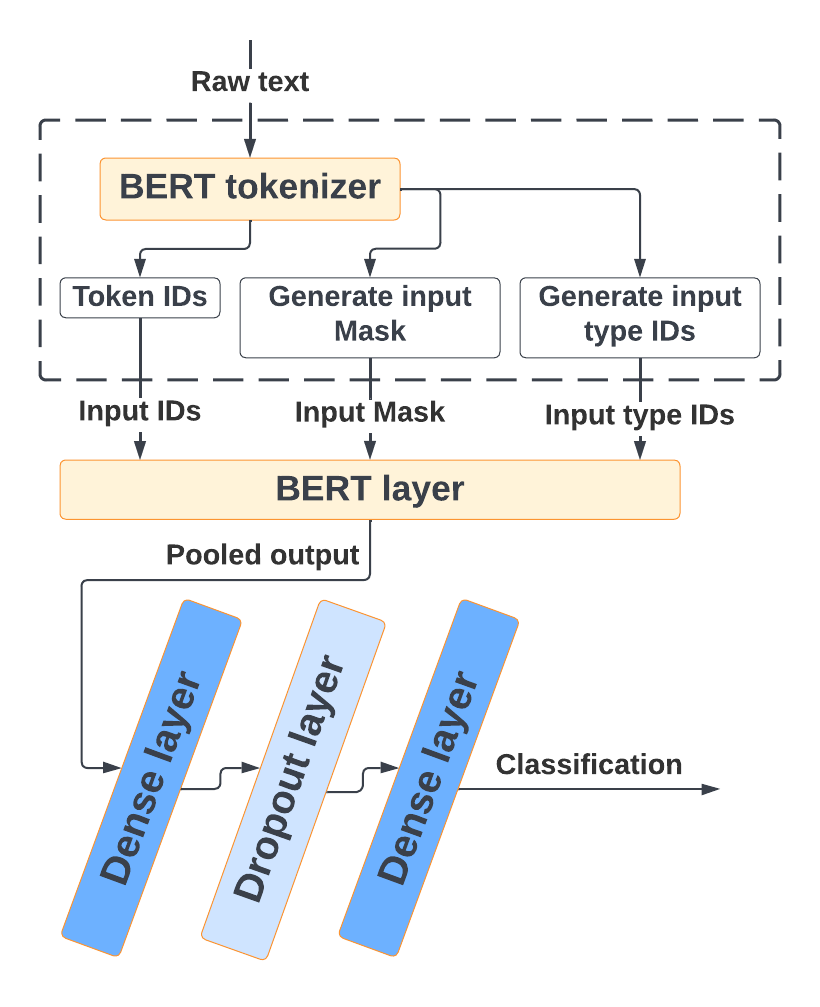}} \label{b}
    \hspace{-1.5\baselineskip}
     \subfigure c.{\includegraphics[width=0.25\textwidth]{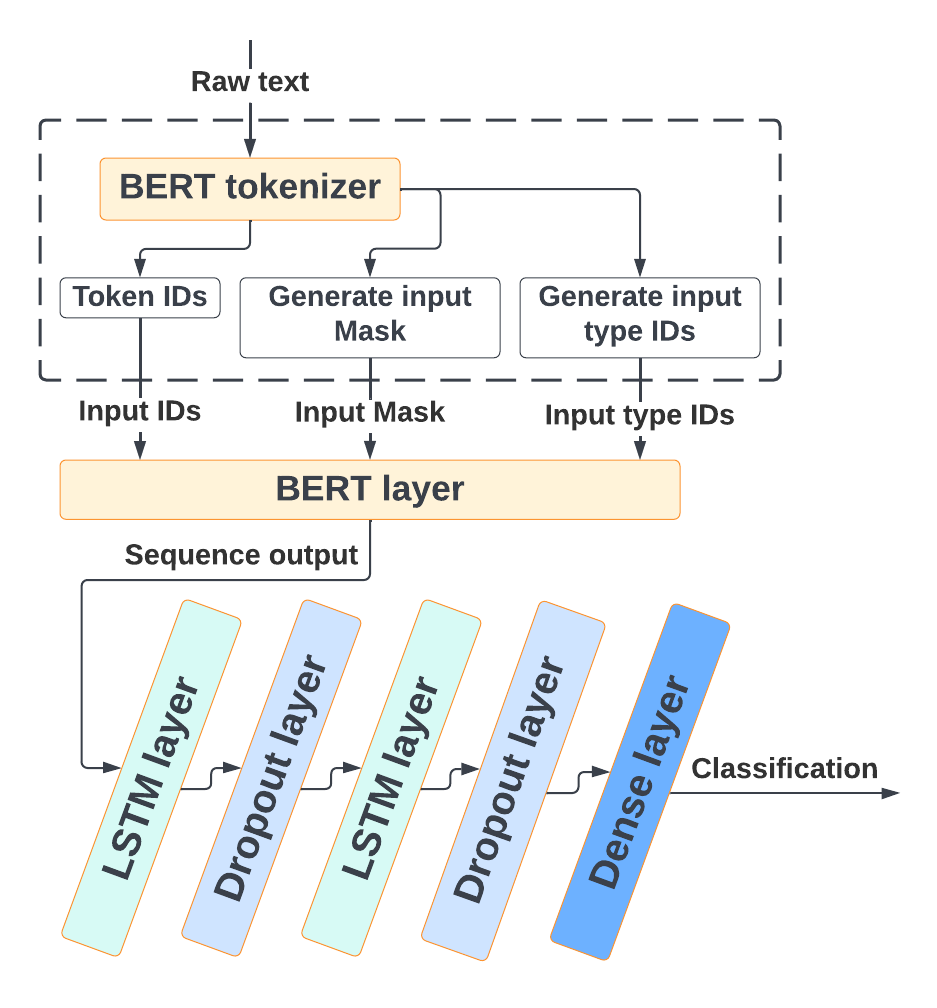}}\label{c}
    \hspace{-2.6\baselineskip}
    \subfigure d.{\includegraphics[width=0.34\textwidth]{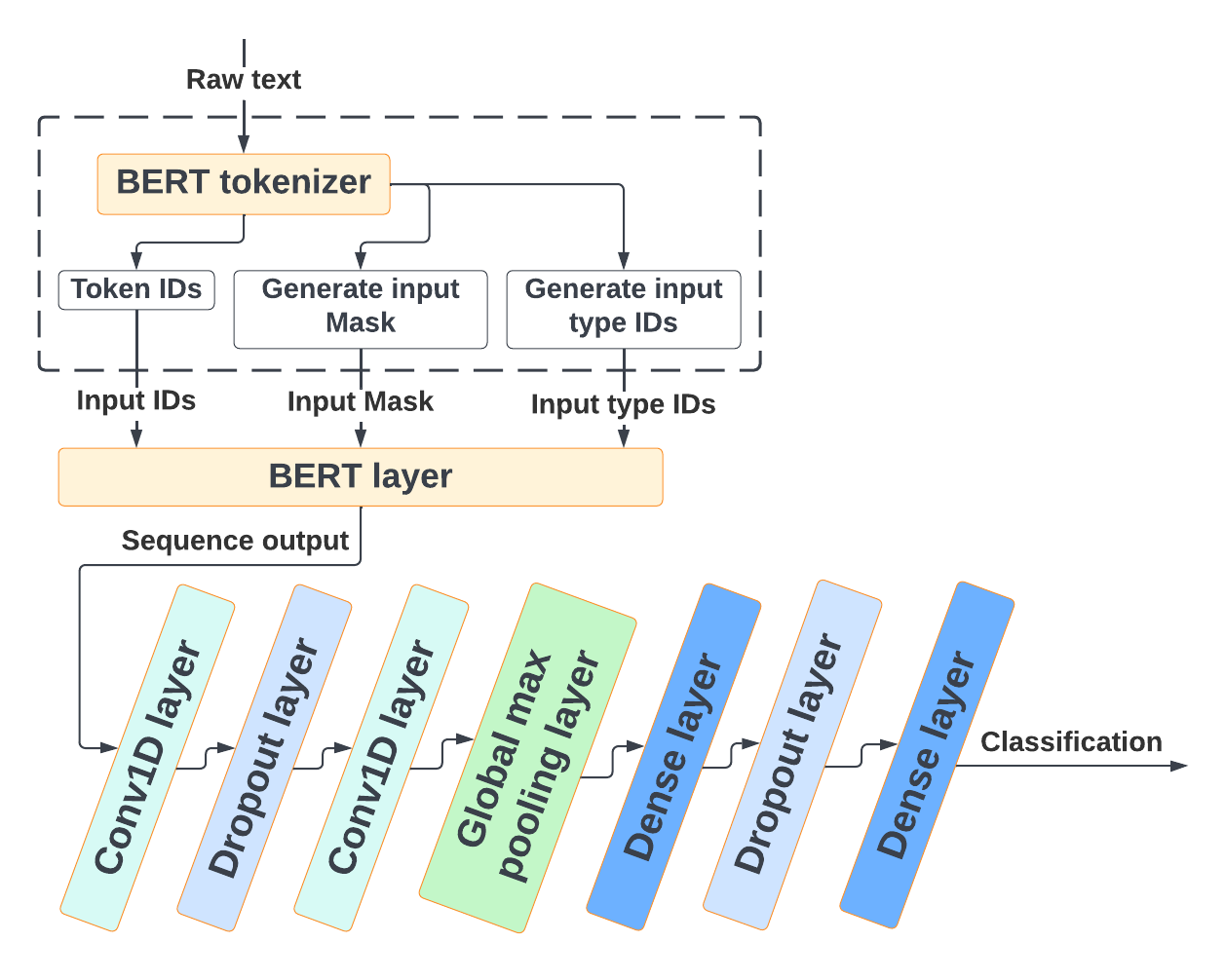}}\label{d}
    \caption{The architectures of integrated deep models: 
    (a.) BERT baseline, (b.) BERT+MLP, (c.) BERT+LSTM, (d.) BERT+CNN }\label{all}
\vspace{-7pt}
\end{figure*}

\vspace{-5pt}
\subsection{Deep models}
\vspace{-7pt}
Trained on large amounts of data \cite{bert}, BERT will certainly lead to excellent performances when using for downstream tasks. It also returns different vectors and contextual embedding for the same word, which extract more information from texts.
In the other hand, deep learning networks provide many advantages for NLP, where CNN and RNN are majorly used for text classification. Therefore,  we implemented 4 models integrating BERT with other popular NN models such as \textbf{MLP}, \textbf{CNN} and \textbf{LSTM}.
At first, we assess the contextual information derived from BERT.
We fine-tune them using our datasets to get its contextual representations and then, ensemble models with several ensemble learning techniques: aggregation and stacking, aiming to improve performance and robustness, and to get better classification. 
The Figure \ref{bert++} shows the general architecture of our models: Text data needs to be transformed to numeric token ids then arranged in several Tensors before being input to BERT-model, here, TensorFlow Hub provides a matching BERT-preprocessor (tokenizer) for each of the BERT models, which implements this transformation using TF.text library\footnote{\url{https://www.tensorflow.org/text/tutorials/classify\_text\_with\_bert}\label{}}. Our Bert-Model you will return 512 dimension embedding for each token: 'sequence output' and 'Pooled output', which, will be fed into the created NN (Keras layer). 

In this work, we used BERT Tensorflow Hub\footnote{\url{https://tfhub.dev/google/collections/bert/1}\label{}} to compute vector-space representations of datasets, to implement 4 different deep models.

\textbf{1) BERT baseline:} (Figure 3.a)
We used BERT uncased L-4 H-512 A-8 model: 4 hidden layers (L), 512 hidden size (output size of 512 dimensions) (H). 
We took the pooled output and integrated a dropout and a dense layer (Softmax activation).
\textbf{2) BERT+MLP:} (Figure 3.b)
MLP is a type of traditional neural networks that are highly adaptable. They consist in 3+ layers of neurons. 2 dense and one dropout layers are added to BERT's pooled output.
\textbf{3) BERT+LSTM} (Figure 3.c):
LSTM is a RNN that update hidden layers using memory cells, and appear to be effective in sequential learning long-range text dependencies. 
We integrated into BERT's sequence output, 2 LSTM layers (512 units) followed by a dropout and a dense layer.
\textbf{4) BERT+CNN} (Figure 3.d):
CNNs are deeper and sparsely connected, allowing to efficiently find patterns in noisy texts. Here, 2 CNN (conv1D with ReLu activation), a Global Max Pooling and 2 dense layers are integrated into BERT's sequence output. 
\vspace{-7pt}
\subsection{Ensemble Learning}
\vspace{-5pt}
Since Neural networks are nonlinear, they have high variance and low bias, being very sensitive to noisy data. Thus, there is no guarantee to exhibit low generalization error when predicting. Hence, we implemented ensemble learning to improve model performance by reducing these issues' effects. 
We used 4 ensembling methods combining BERT+MLP, BERT+CNN and BERT+LSTM, with stacking and aggregation:
\textbf{1) Soft Voting} Or averaging, merges several fine tuned models trained on the same dataset. 
We took the average of predicted class probabilities of each individual classifier $C_{j}$ and then, used argmax to obtain the final class as shown in equation  \ref{eq:softvoting}. The predicted probabilities were treated equally ($\omega_{j}=1$ for each $\textit{j}th$ model).
\vspace{-7pt}
\begin{align}
\hat{y} = argmax_{i\in\left \{ 1,...c \right \} j\in\left \{ 1,...m \right \} }  \omega_{j}\hat{p}_{ij}
\label{eq:softvoting}
\end{align}
$i$ is the class value in data and the class probability $\hat{p}$ is
$\forall i$, $\hat{p}\left ( y=i\right ) = \sum_{j=1}^{m} \frac{ \hat{p}\left ( y=i |C_{j} \right )}{m}.$\\ 

\vspace{-7pt}
\begin{figure}
    \includegraphics[width=.4\textwidth]{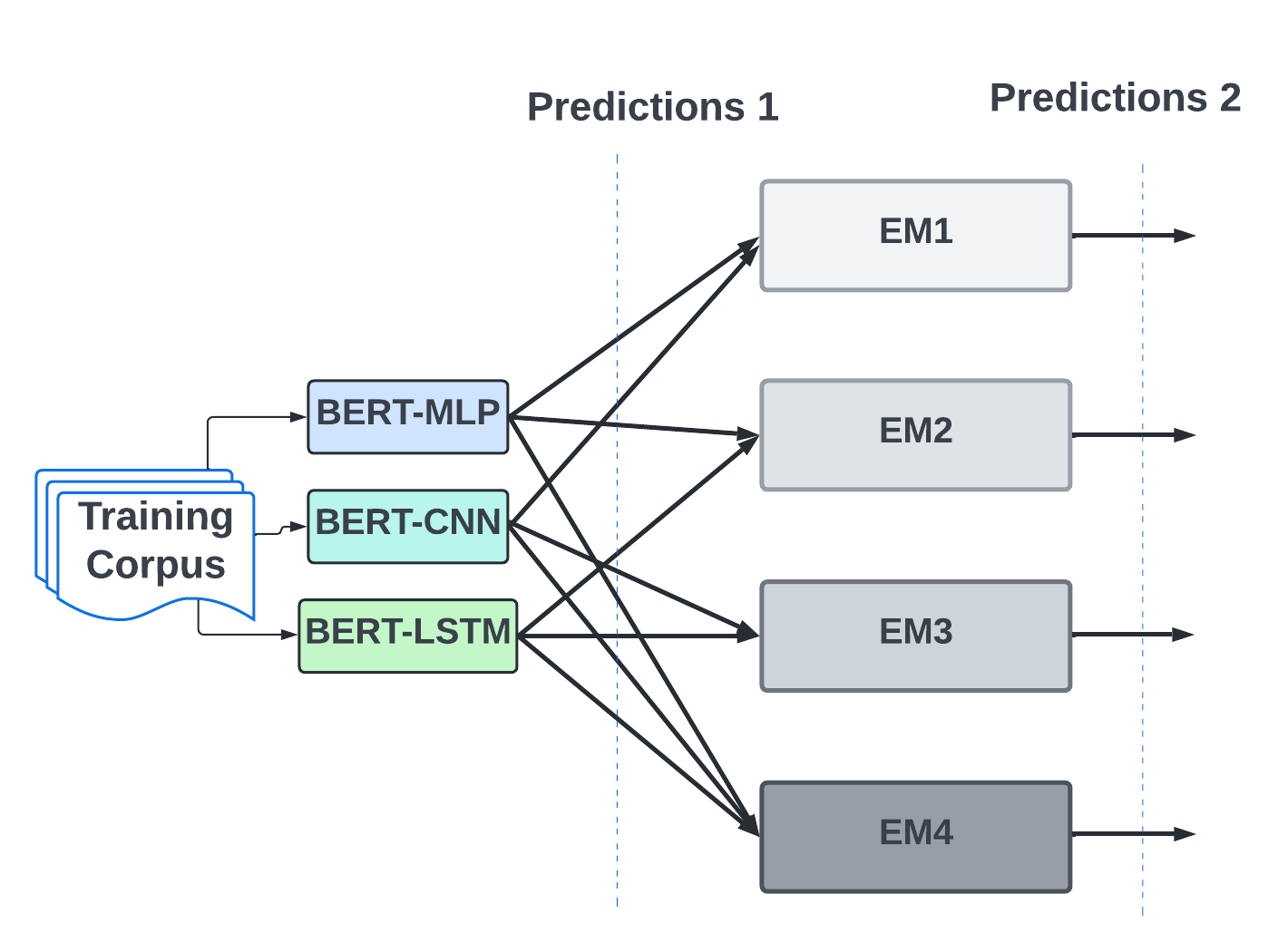}
    \caption{Different \textbf{E}nsemble learning \textbf{M}odels (\textbf{EM}):\\ \textbf{EM1}: BERT-MLP + BERT-CNN, \textbf{EM2:} BERT-MLP + BERT-LSTM \textbf{EM3:}, BERT-CNN + BERT+LSTM \textbf{EM4:} and BERT-MLP + BERT-CNN + BERT-LSTM}\label{ensembles}
    \vspace{-9pt}
\end{figure}

\textbf{2) Maximum voting }
Consider the maximum prediction probability from the models.
We have $C(x) = \hat{p}$: the probability distribution over the c classes $y$ where:\\ $\hat{p} = \left \{ \hat{p}(y=0|C), \hat{p}(y=1|C),... \hat{p}(y=m|C) \right \}$. 
Thus, the max value ensemble result is the class with the maximum probability among the classifiers, as illustrated in the equation \ref{eq:maxvalue}.
\vspace{-7pt}
\begin{align}
\hat{y} = argmax  _ {i\in\left \{ 1,...c \right \} } \hat{p}\left ( y=i| C \right )
\label{eq:maxvalue}
\end{align}
\textbf{3) Hard voting }
Employs the principle of majority voting (of an odd number of classifiers), it takes the predictions of each model and output the most frequent class. We predict the class $\hat{y}$ via majority vote of each of the $m$ classifiers as shown in equation \ref{eq:hardvoting}.
\vspace{-7pt}
\begin{align}
\hat{y} = mode\left \{C_{1}, C_{2},... C_{m} \right \}.
\label{eq:hardvoting}
\end{align}

\textbf{4) Stacked Generalization ensemble}
Or stacking, is an integration strategy \cite{stacking}. It frequently combines heterogeneous trained base learners by training a \textit{meta-model} to output a prediction based on the predictions of the base models. We have implemented the Stratified k-fold cross-validation technique to partition the training set between the models, keeping the same ratio of 10\% for the validation set.
Given the training set $D = \left \{ D_{1} ,  D_{2} ,  D_{3} \right \}$ where $D_{j}$ and $\bar{D_{j}} = D\setminus D_{j}$ are respectively the training set and its corresponding test set, each base $t^{th}$ learner $h_{t}^{(j)}$ is trained on a $D_{j}$, then, their predictions z (equation \ref{eq:stack}) are fed as training set into the meta learner (Linear Regression classifier in our case) which will predict the final predictions.
\vspace{-7pt}
\begin{align}
z =stack\left \{ h_{t}\left ( D \right ) \right \}.
\label{eq:stack}
\vspace{-10pt}
\end{align}

All ensemble models are depicted in Figure \ref{ensembles} which presents the architecture of the integration of multiple deep learning models to form 10 different ensemble models. 
These ensembles have 2 prediction levels: \textit{Predictions1}: the predictions of each model and \textit{Predictions2}: the output after ensembling them using different scenarios.
EM1 is the ensemble of BERT-MLP and BERT-CNN, based on these 2 levels. As Figure \ref{ensembles}  depicts, EM2, EM3 and EM4 are implemented by the same technique.
\vspace{-5pt}

\section{Experiments and Results}
\begin{table*}[]
\centering
\scriptsize
\begin{tabular}{cclcccccc}
\cline{4-9}
\multicolumn{2}{l}{}                                                                                                             & \multicolumn{1}{l|}{}                & \multicolumn{3}{c|}{\textbf{Davidson Dataset}}                                                                                                                                          & \multicolumn{3}{c|}{\textbf{Merged Dataset}}                                                                                                                                            \\ \cline{3-9} 
\multicolumn{2}{l}{}             & \multicolumn{1}{c|}{\textbf{Model}}  & \multicolumn{1}{c|}{\textbf{Accuracy}}                      & \multicolumn{1}{c|}{\textbf{F1 Score}}                      & \multicolumn{1}{c|}{\textbf{Precision}}                     & \multicolumn{1}{c|}{\textbf{Accuracy}}                      & \multicolumn{1}{c|}{\textbf{F1 Score}}                      & \multicolumn{1}{c|}{\textbf{Precision}}                     \\ \cline{3-9} 
\multicolumn{2}{l|}{}                                                                                                            & \multicolumn{1}{l|}{BERT (baseline)} & \multicolumn{1}{c|}{{\color[HTML]{212121} 0.8498}}          & \multicolumn{1}{c|}{{\color[HTML]{212121} 0.7209}}          & \multicolumn{1}{l|}{{\color[HTML]{212121} 0.6894}}          & \multicolumn{1}{c|}{{\color[HTML]{212121} 0.79164}}         & \multicolumn{1}{c|}{{\color[HTML]{212121} 0.7571}}          & \multicolumn{1}{c|}{{\color[HTML]{212121} 0.7466}}          \\ \cline{3-9} 
\multicolumn{2}{l|}{}                                                                                                            & \multicolumn{1}{l|}{BERT-MLP}        & \multicolumn{1}{c|}{{\color[HTML]{212121} 0.876}}           & \multicolumn{1}{c|}{{\color[HTML]{212121} 0.7401}}          & \multicolumn{1}{c|}{{\color[HTML]{212121} 0.7209}}          & \multicolumn{1}{c|}{{\color[HTML]{212121} \textbf{\textcolor{blue}{0.8087}}}} & \multicolumn{1}{c|}{{\color[HTML]{212121} \textbf{\textcolor{blue}{0.7704}}}} & \multicolumn{1}{c|}{{\color[HTML]{212121} \textbf{\textcolor{blue}{0.764}}}}  \\ \cline{3-9} 
\multicolumn{2}{l|}{}                                                                                                            & \multicolumn{1}{l|}{BERT-CNN}        & \multicolumn{1}{c|}{{\color[HTML]{212121} 0.8943}}          & \multicolumn{1}{c|}{{\color[HTML]{212121} 0.7355}}          & \multicolumn{1}{c|}{{\color[HTML]{212121} 0.7379}}          & \multicolumn{1}{c|}{{\color[HTML]{212121} 0.7853}}          & \multicolumn{1}{c|}{{\color[HTML]{212121} 0.74902}}         & \multicolumn{1}{c|}{{\color[HTML]{212121} 0.7391}}          \\ \cline{3-9} 
\multicolumn{2}{l|}{}                                                                                                            & \multicolumn{1}{l|}{BERT-LSTM}       & \multicolumn{1}{c|}{{\color[HTML]{212121} \textbf{\textcolor{blue}{0.8968}}}} & \multicolumn{1}{c|}{{\color[HTML]{212121} \textbf{\textcolor{blue}{0.7548}}}} & \multicolumn{1}{c|}{{\color[HTML]{212121} \textbf{\textcolor{blue}{0.7397}}}} & \multicolumn{1}{c|}{{\color[HTML]{212121} 0.8013}}          & \multicolumn{1}{c|}{{\color[HTML]{212121} 0.762}}           & \multicolumn{1}{c|}{{\color[HTML]{212121} 0.7559}}          \\ \cline{3-9} 
\multicolumn{1}{l}{}                                               & \multicolumn{1}{l}{\textbf{}}                               & \textbf{}                            & \multicolumn{1}{l}{}                                        & \multicolumn{1}{l}{}                                        & \multicolumn{1}{l}{}                                        & \multicolumn{1}{l}{}                                        & \multicolumn{1}{l}{}                                        & \multicolumn{1}{l}{}                                        \\ \hline
\multicolumn{1}{|c|}{}                                             & \multicolumn{1}{c|}{}                                       & \multicolumn{1}{l|}{EM1}             & \multicolumn{1}{c|}{{\color[HTML]{212121} 0.9005}}          & \multicolumn{1}{c|}{{\color[HTML]{212121} 0.7558}}          & \multicolumn{1}{c|}{{\color[HTML]{212121} 0.7468}}          & \multicolumn{1}{c|}{{\color[HTML]{212121} 0.8045}}          & \multicolumn{1}{c|}{{\color[HTML]{212121} 0.7672}}          & \multicolumn{1}{c|}{{\color[HTML]{212121} 0.7586}}          \\ \cline{3-9} 
\multicolumn{1}{|c|}{}                                             & \multicolumn{1}{c|}{}                                       & \multicolumn{1}{l|}{EM2}             & \multicolumn{1}{c|}{{\color[HTML]{212121} 0.8943}}          & \multicolumn{1}{c|}{{\color[HTML]{212121} 0.7554}}          & \multicolumn{1}{c|}{{\color[HTML]{212121} 0.7354}}          & \multicolumn{1}{c|}{{\color[HTML]{212121} \textbf{\textcolor{blue}{0.8158}}}} & \multicolumn{1}{c|}{{\color[HTML]{212121} \textbf{\textcolor{blue}{0.7783}}}} & \multicolumn{1}{c|}{{\color[HTML]{212121} \textbf{\textcolor{blue}{0.7724}}}} \\ \cline{3-9} 
\multicolumn{1}{|c|}{}                                             & \multicolumn{1}{c|}{}                                       & \multicolumn{1}{l|}{EM3}             & \multicolumn{1}{c|}{{\color[HTML]{212121} \textbf{\textcolor{red}{0.9047}}}} & \multicolumn{1}{c|}{{\color[HTML]{212121} \textbf{\textcolor{blue}{0.7596}}}} & \multicolumn{1}{c|}{{\color[HTML]{212121} \textbf{\textcolor{blue}{0.7519}}}} & \multicolumn{1}{c|}{{\color[HTML]{212121} 0.8091}}          & \multicolumn{1}{c|}{{\color[HTML]{212121} 0.7734}}          & \multicolumn{1}{c|}{{\color[HTML]{212121} 0.7641}}          \\ \cline{3-9} 
\multicolumn{1}{|c|}{}                                             & \multicolumn{1}{c|}{\textbf{Soft voting}} & \multicolumn{1}{l|}{EM4}             & \multicolumn{1}{c|}{{\color[HTML]{212121} 0.9018}}          & \multicolumn{1}{c|}{{\color[HTML]{212121} 0.7589}}          & \multicolumn{1}{c|}{{\color[HTML]{212121} 0.7469}}          & \multicolumn{1}{c|}{{\color[HTML]{212121} 0.812}}           & \multicolumn{1}{c|}{{\color[HTML]{212121} 0.7746}}          & \multicolumn{1}{c|}{{\color[HTML]{212121} 0.7669}}          \\ \cline{2-9} 
\multicolumn{1}{|c|}{}                                             & \multicolumn{1}{c|}{}                                       & \multicolumn{1}{l|}{EM1}             & \multicolumn{1}{c|}{{\color[HTML]{212121} 0.9009}}          & \multicolumn{1}{c|}{{\color[HTML]{212121} 0.7564}}          & \multicolumn{1}{c|}{{\color[HTML]{212121} 0.747}}           & \multicolumn{1}{c|}{{\color[HTML]{212121} 0.8077}}          & \multicolumn{1}{c|}{{\color[HTML]{212121} 0.7715}}          & \multicolumn{1}{c|}{{\color[HTML]{212121} 0.7622}}          \\ \cline{3-9} 
\multicolumn{1}{|c|}{}                                             & \multicolumn{1}{c|}{}                                       & \multicolumn{1}{l|}{EM2}             & \multicolumn{1}{c|}{{\color[HTML]{212121} 0.8939}}          & \multicolumn{1}{c|}{{\color[HTML]{212121} 0.7549}}          & \multicolumn{1}{c|}{{\color[HTML]{212121} 0.7348}}          & \multicolumn{1}{c|}{{\color[HTML]{212121} \textbf{\textcolor{red}{0.8162}}}} & \multicolumn{1}{c|}{{\color[HTML]{212121} \textbf{\textcolor{blue}{0.7786}}}} & \multicolumn{1}{c|}{{\color[HTML]{212121} \textbf{\textcolor{blue}{0.7726}}}} \\ \cline{3-9} 
\multicolumn{1}{|c|}{}                                             & \multicolumn{1}{c|}{}                                       & \multicolumn{1}{l|}{EM3}             & \multicolumn{1}{c|}{{\color[HTML]{212121} \textbf{\textcolor{blue}{0.9034}}}} & \multicolumn{1}{c|}{{\color[HTML]{212121} 0.7566}}          & \multicolumn{1}{c|}{{\color[HTML]{212121} \textbf{\textcolor{blue}{0.7494}}}} & \multicolumn{1}{c|}{{\color[HTML]{212121} 0.8077}}          & \multicolumn{1}{c|}{{\color[HTML]{212121} 0.7715}}          & \multicolumn{1}{c|}{{\color[HTML]{212121} 0.7622}}          \\ \cline{3-9} 
\multicolumn{1}{|c|}{}                                             & \multicolumn{1}{c|}{\textbf{Max value}}  & \multicolumn{1}{l|}{EM4}             & \multicolumn{1}{c|}{{\color[HTML]{212121} 0.9026}}          & \multicolumn{1}{c|}{{\color[HTML]{212121} \textbf{\textcolor{blue}{0.7588}}}} & \multicolumn{1}{c|}{{\color[HTML]{212121} 0.748}}           & \multicolumn{1}{c|}{{\color[HTML]{212121} 0.814}}           & \multicolumn{1}{c|}{{\color[HTML]{212121} 0.7773}}          & \multicolumn{1}{c|}{{\color[HTML]{212121} 0.769}}           \\ \cline{2-9} 
\multicolumn{1}{|c|}{}                                             & \multicolumn{1}{c|}{\textbf{Hard voting}}                   & \multicolumn{1}{l|}{EM4}             & \multicolumn{1}{c|}{{\color[HTML]{212121} 0.9001}}          & \multicolumn{1}{c|}{{\color[HTML]{212121} \textbf{0.7595}}} & \multicolumn{1}{c|}{{\color[HTML]{212121} 0.7457}}          & \multicolumn{1}{c|}{{\color[HTML]{212121} 0.8085}}          & \multicolumn{1}{c|}{{\color[HTML]{212121} 0.771}}           & \multicolumn{1}{c|}{{\color[HTML]{212121} 0.7628}}          \\ \cline{2-9} 
\multicolumn{1}{|c|}{\textbf{Ensembler Method}} & \multicolumn{1}{c|}{\textbf{Stacking}}                      & \multicolumn{1}{l|}{EM4}             & \multicolumn{1}{c|}{{\color[HTML]{212121} 0.776 }}                & \multicolumn{1}{c|}{{\color[HTML]{212121} \textbf{\textcolor{red}{0.9706}} }}                & \multicolumn{1}{c|}{{\color[HTML]{212121} \textbf{\textcolor{red}{0.9430 }}}}                & \multicolumn{1}{c|}{{\color[HTML]{212121} 0.463}}          & \multicolumn{1}{c|}{{\color[HTML]{212121} \textbf{\textcolor{red}{0.9278}}}}          & \multicolumn{1}{c|}{{\color[HTML]{212121} \textbf{\textcolor{red}{0.8654}}}}          \\ \hline
\end{tabular}
\label{table:finalresults}
\vspace{0.15cm}
\caption{Experimental Results for the Davidson dataset and DHO dataset. }
\vspace{-7pt}
\end{table*}

\subsection{Data Pre-processing}
\vspace{-5pt}
We started pre-processing our raw data:
1) Switch tweets to lower case, 2) Delete URLs, 3) Remove users names, 4) Shorten prolonged words ("yeeessss" to "yes"...), 5) Keep stop words, 6) Remove punctuation marks, unknown uni-codes, and additional delimiting characters, 7) Remove hashtags (\#) and correct their texts (e.g, "\#notracism" to "not racism"), 8) Eliminate tweets of length less than 2, and 9) Remove emoticons. 
\vspace{-7pt}
\subsection{Data analysis platform and evaluation metrics}
\vspace{-5pt}
The implemented models are trained with various fine-tuning strategies (batch size of 32 for 50 epochs) on Colab. We created our custom optimizer with a learning rate of 2e-5, and experimented with AdamW optimizer and Sparse Categorical Cross-entropy loss functions.
In order to utilize features from each label, we introduced weights in the training phase. 
Thus, classifiers performances are measured via macro averaged F1 score, accuracy, precision and recall scores.
\vspace{-7pt}
\subsection{Results and Interpretations}
\vspace{-5pt}
The results are shown in Table 1 and the best results of each group of the same ensemble and are highlighted in blue and the best overall ones are in red.
We set up early stopping with Validation Accuracy as a monitor parameter to prevent overfitting.
Following experiments are mainly based on \textit{Davidson} dataset and \textit{DHO} dataset.
\textbf{Davidson dataset:}
As illustrated in Table 1, BERT+LSTM has shown the best results on the test set.
Moreover, all the models outperform BERT baseline, which prove the importance of adding NN classifiers for such a task. 
As for the aggregation ensembles, all the approaches outperformed single models, it shows obviously better results, especially the Soft Voting of BERT+LSTM with BERT+CNN,
as well as Hard Voting ensembling (in bold) that outperformed both of the other aggregation ensembles. \textbf{DHO dataset:}
Models trained and tested on this dataset gave better performance than on Davidson's. Getting the most performed model BERT+MLP.
Moreover, aggregation ensembles outperformed each of these single models, getting the best result when ensembling the 2 most performed models: BERT+MLP and BERT+LSTM. All of the 3 ensemble approaches have given almost similar results, but we can observe that Maximum value and Hard Voting are the best ones.
Unlike BERT-CNN, BERT-MLP and BERT-LSTM gave the best performance on DHO and Davidson datasets respectively, this is related to the type of corpora and the subject we are dealing with: NLP Twitter text classification with small imbalanced dataset. As for the stacking approach, it outperformed all the other ensembles in both datasets, giving highest F1 and precision, even with the lower accuracy, which is not always guaranteed in this ensemble and unexpectedly, Davidson accuracy results were better then DHO's. We refer these results to the  meta classifier used (Linear Regression), which is not a complex multi layered model that can correctly deal with the used datasets. And, as David H. et al. stated \cite{stacking}, for almost any real-world generalization problem one should use some version of stacked generalization to minimize the generalization error rate.
Moreover, the main reason for getting lower accuracy is not easy to be interpretable since BERT (as any other transformer) is a black box, so it is not easy to understand its functioning. Meanwhile, we refer also these results to the imbalanced corpora used in our experiments.
Figure 5 shows the confusion matrix to present clearly the ensembles' classification performance of the best performed models: soft voting EM3 for Davidson, and Max value EM2 for DHO. The classification performance of both models is very good for each class, especially for 'Offensive' and 'Neither' (Highest True Positives) since they represent the biggest percentages of each dataset. Moreover, 'Hateful' class classification error rate is decreased in DHO because this corpora is more balanced.
\vspace{-5pt}
\begin{figure}
    \centering
    \subfigure a.{\includegraphics[width=0.23\textwidth]{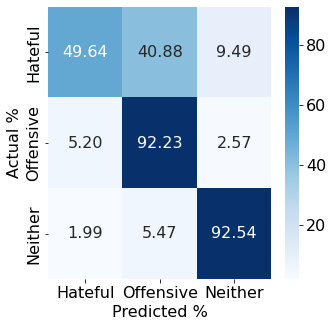}}\label{a}
    \hspace{-1\baselineskip}
    \subfigure b.{\includegraphics[width=0.23\textwidth]{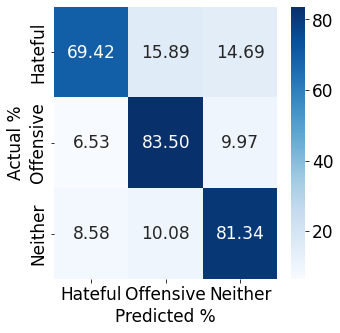}} \label{b}
    \caption{Confusion matrix of ensembles: 
    (a.) Soft voting EM3 - Davidson, (b.) Max value EM2 - DHO }\label{confusion}
\end{figure}

\section{Discussions and future work}
\vspace{-5pt}
\subsection{Discussions and Challenges}
\begin{figure}
    \includegraphics[width=.43\textwidth]{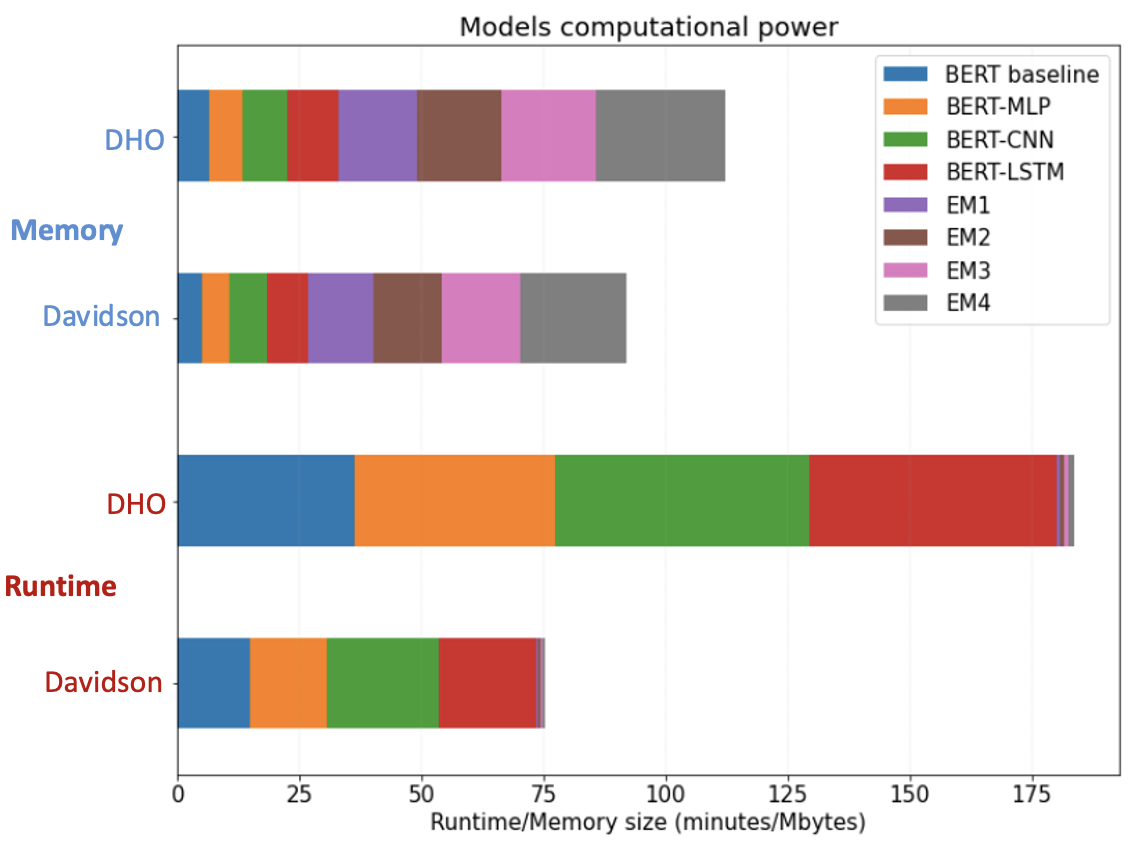}
    \caption{Runtime and memory models.}\label{computer}
    \vspace{-7pt}
\end{figure}

\vspace{-5pt}
\textbf{Data:} As shown in the section above, we worked on a small-size  imbalanced datasets, where there is a huge gap between class distribution (in Davidson dataset ‘Hateful’ label takes 5.77\% while ‘Offensive’ takes 77.34\% of all data) Figure \ref{data}.
This led to many issues we faced during training such as the overfitting problems.
\textbf{Computational power:} Our proposed approaches require a strong GPU, and even with the use of Google Colab Pro for training, we were restricted to some limitations (Number of BERT and deep learning layers). We ended up getting a 28,765,188-parameter BERT baseline, 29,027,844-parameter BERT-MLP, 28,835,428-parameter BERT-CNN and  32,963,588-parameter BERT-LSTM, which, increases the models' complexity, thus, required more computational resources as displayed in Figure \ref{computer}.

Obviously, DHO require more resources than Davidson.
And even getting less number of parameters, training BERT-CNN took more memory size than BERT-MLP. Although requiring more memory, the ensembles were very faster than baselines' training, which explains their efficiency. The best least resouces consuming models are EM1 and EM3 for both datasets.
\textbf{Multi-label classification:} We worked on multi-label datasets, which add more complexity to the classifications. When compared with the previous works, our results didn’t often overcome the binary classifications on the same datasets.
\textbf{Stacking ensemble:}
Dealing with BERT+NN, we couldn't implement directly the pre-defined stacking ensemble functions of Sklearn library. Thus, we manually elaborated the required functions (Defining: stacked dataset got from the predictions of base learners, the meta learner and, its stacked final predictions). 
\vspace{-7pt}
\subsection{Future work} 
\vspace{-6pt}
Various improvements can be added to our study in the future to enhance hate speech detection approach with ensemble transfer learning. It starts with the data where we can work more on the labels’ distribution. Taking into consideration balance status, we can use several techniques like data augmentation. Dealing with the dataset bias issue and the granularity of hate speech content, we can get benefit further from K-BERT (using Knowledge Graph) \cite{kbert}, it helps also to handle the issue of the data scarcity, especially in other languages, thus, detecting hateful and offensive content from unlabeled corpora. Adding to that, we can check our models' generalization by testing them on other datasets, from different social media resources like Facebook etc. As for the models' implementation, we can use TensorFlow Hub BERT architecture similar to BERT base Hugging Face model’s (L=12, H=768 and A=12), or at least, increase the number of parameters and test their impact on models' performances. Moreover, we can employ other ensemble learning techniques and enhance the stacking ensemble we built, we can also improve the baseline models and build new and more complex deep learning classifiers and combine them with other transformers than BERT and compare their effects.
\vspace{-7pt}
\section{Conclusion}
For the purpose of hate speech detection on social media, this paper worked on several public datasets to build large more balanced corpora used to train and test transfer learning classifiers. Combining BERT transformer with several deep neural networks, we managed to get heterogeneous multi-label classifiers that successfully detect hateful and offensive content from Twitter. We, then , improved their performance using different aggregating and stacking ensemble techniques, the latter significantly outperformed the baselines and the other ensembles' predictions even with giving low accuracy, which is a future subject to work on, where we aim to develop more powerful and resilient stacking meta-classifiers.
\vspace{-7pt}

\end{document}